\documentclass[manuscript]{acmart}
\setcopyright{none}
\usepackage{hyperref}
\usepackage{url}
\usepackage{algorithm}
\usepackage{algpseudocode}
\usepackage{booktabs}
\usepackage{graphicx}
\usepackage{bm}
\usepackage{float}            
\usepackage{algpseudocode}   

\usepackage{amsmath, amsfonts, amssymb}

\title{Generative human motion mimicking through feature extraction in denoising diffusion settings}

\author{Alexander Okupnik}
\affiliation{
  \department{Department of Computer Science and Information Systems}
  \institution{University of Liechtenstein}
  \city{Vaduz}
  \country{Liechtenstein}
}
\email{alexander.okupnik@uni.li}

\author{Johannes Schneider}
\affiliation{
  \department{Department of Computer Science and Information Systems}
  \institution{University of Liechtenstein}
  \city{Vaduz}
  \country{Liechtenstein}
}
\email{johannes.schneider@uni.li}

\author{Kyriakos Flouris}
\affiliation{
  \institution{University of Cambridge}
  \city{Cambridge}
  \country{United Kingdom}
}
\email{kyriakos.flouris@mrc-bsu.cam.ac.uk}

\begin{document}

\begin{abstract}
Recent success with large language models has sparked a new wave of verbal human-AI interaction. While such models support users in a variety of creative tasks, they lack the embodied nature of human interaction. Dance, as a primal form of human expression, is predestined to complement this experience. To explore creative human-AI interaction exemplified by dance, we build an interactive model based on motion capture (MoCap) data. It generates an artificial other by partially mimicking and also "creatively" enhancing an incoming sequence of movement data. It is the first model, which leverages single-person motion data and high level features in order to do so and, thus, it does not rely on low level human-human interaction data. It combines ideas of two diffusion models, motion inpainting, and motion style transfer to generate movement representations that are both temporally coherent and responsive to a chosen movement reference. The success of the model is demonstrated by quantitatively assessing the convergence of the feature distribution of the generated samples and the test set which serves as simulating the human performer. We show that our generations are first steps to creative dancing with AI as they are both diverse showing various deviations from the human partner while appearing realistic.
\end{abstract}

\maketitle

\section{Introduction}

\begin{figure}
    \vspace{-6pt}
    \centering
    \includegraphics[width=0.85\linewidth]{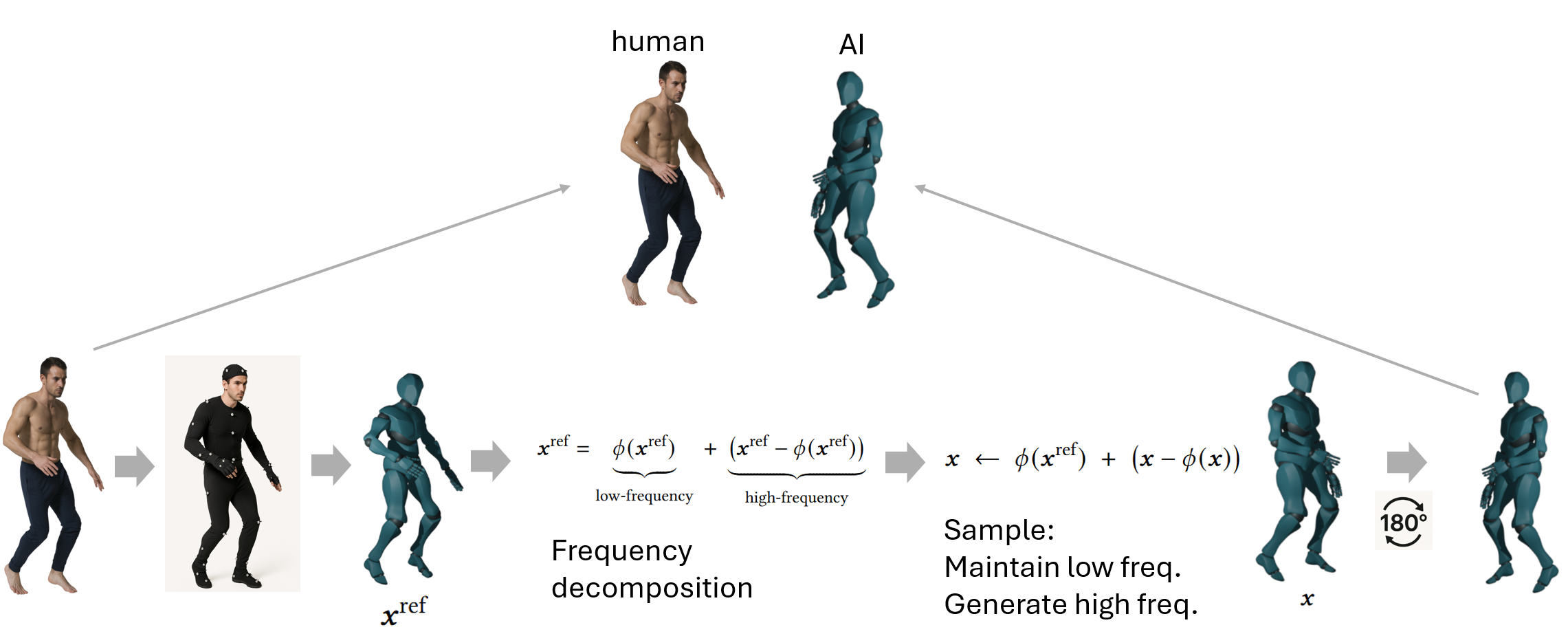}
    \vspace{-6pt}
    \caption{Human dances with AI in 3D. 3D motion data is captured (e.g. using a motion tracking suit) and decomposed into low and high frequency components. Low frequency components are maintained to align AI and human movements and create a form of interactivity, while high frequency components are sampled from a diffusion model combining multiple ideas to express diversity.}
    \label{fig:overview}
    \vspace{-6pt}
\end{figure}

Across many cultures and millennia, dance has remained a vital element of the social fabric—as self-expression, ritual, social participation, and, in professional contexts, artistic practice. As a nonverbal mode of interaction, dance also opens a new dimension for human–AI collaboration beyond the recent successes of Large Language Models (e.g., ChatGPT), where interaction is mostly textual or visual. In music, for example, Omax learns stylistic patterns from a performer in real time and improvises with them by extracting high-level features (e.g., pitch, beat, dynamics) and recombining them meaningfully \cite{omax}. This points to promising avenues for nonverbal, improvisational co-creation with machines, where the system adapts fluidly to a human partner rather than executing a fixed script.

However, bringing this vision to dance introduces distinctive challenges. Human movement can be captured within different technological settings, e.g., motion-capture suits with inertial sensors measuring acceleration or cameras capturing markers. The most recent work in dance generation emphasizes single-person music-to-movement synthesis, where a model generates motion conditioned on audio \cite{siyao2022bailando3ddancegeneration,liu2025dgfmbodydancegeneration,huang2023dancerevolutionlongtermdance,Wang_2023}.  Although the multimodal line of research has produced compelling results, it addresses a different problem: mapping sound to motion rather than building an interactive loop between a human dancer and an adaptive AI partner. To study nonverbal interaction directly, several datasets of duet dances. However, models trained on such duet data are optimized to replicate human–human coupling. Their learned dynamics may transfer poorly to human–AI scenarios—especially in social or improvisational settings (e.g., salsa) where movement departs from stereotyped patterns and timing is negotiated moment-to-moment, or in an interactive partner dance in Figure \ref{fig:overview}.

An additional complication is performer awareness. The very knowledge that one is interacting with a machine-generated avatar can change behavior, pushing the system into out-of-distribution regimes relative to its training data, leading to poorly generated movements that deviate strongly from potential real-world behavior. In a performative study of an interactive model \cite{10.1007/978-3-319-77583-8_17}, the relationship between performer and avatar shifted once the performer realized that the system was learning sequences on the fly, altering both expectation and response. This observation echoes findings in human–computer interaction more broadly: transparency and perceived agency feed back into user behavior. For dance, where timing, tension, and attention are coupled, even subtle changes in intention can cascade into different motion distributions—precisely those that duet-trained models may not anticipate and therefore do not perform well in such a setup in contrast to our work exhibits greater flexibility through the use of high-level features.

To address these gaps, we propose simulating an interaction between a human performer and an AI avatar using only single-person motion data. Rather than learning pairwise coordination directly from duet datasets, we leverage high-level features extracted from solo sequences to steer an interactive generative model. Our underlying rationale and understanding of dancing is that partners need a form of alignment, but also need to differ in their motion to achieve a playful, creative interaction. We implement this idea by letting the AI mimic low frequency movements of the human partner and allowing it more freedom in high frequency movements (as illustrated in Figure \ref{fig:overview}.
As our backbone, we adopt a modern, generative dance architecture EDGE \cite{tseng2023edge}, selected for its strong motion quality and its motion-inpainting capability, which supports on-the-fly, time-coherent extensions of partially observed movement. On top of EDGE, we integrate Iterative Latent Variable Refinement (ILVR) \cite{choi2021ilvrconditioningmethoddenoising} to edit the incoming sequence towards the style of a model-generated reference sample. Conceptually, ILVR provides a continuous “style pull”: by iteratively refining the latent variables, the system can nudge the generated motion toward the target stylistic attributes while respecting the performer’s evolving input.

This design offers three advantages. First, it removes the dependence on duet datasets, which are costly to collect and may hard-code human–human coordination patterns that do not generalize to a human–AI setting. To generate pair dancing sequences, traditional models required both partner trajectories, since they were primarily trained to simply ``copy'' the trajectory of one of the partners. In contrast, in our model, we need only one partner and follow the low-level frequency components of his or her movements but allow for creativity by altering his or her high-frequency components. Second, it offers options to regulate the interaction of the model with the incoming sequence during inference time. Third, it supports improvisation: because EDGE can extend motion as it arrives, the system can respond to unexpected phrasing or novel transitions, rather than forcing the interaction into pre-learned motifs.

A key parameter in this mimicry is the degree of similarity between generated motion and the chosen reference. We empirically demonstrate that increasing the number of ILVR refinement steps monotonically raises a set of similarity metrics between the generated and reference samples, effectively providing a tunable “follow strength.” Importantly, this control does not collapse diversity: beyond a threshold, additional refinement returns diminishing gains in similarity while preserving responsiveness to new performer cues, indicating a practical operating region for interactive use.

In summary, we contribute (i) an interaction framework for human–AI dance that relies on solo data and feature-level guidance rather than duet training, (ii) an integration of EDGE with ILVR for real-time, style-conditioned motion editing, and (iii) an analysis of similarity as a controllable parameter governing the balance between stylistic alignment and improvisational freedom of the generative model. Together, these components move toward AI dance partners that can listen to movement, modulate its style and form an interaction with a human performer.
  
\section{Related Work}

\paragraph{Dance Generation}
In early works using deep learning techniques for dance generation, many authors used Recurrent Neural Networks (RNNs) as the fundamental architecture given their success in modeling time series data. \cite{graves2014generatingsequencesrecurrentneural} combined them with long short-term memory (LSTM) in order to enhance the models capability for store information along longer sequences.
\cite{crnkovicfriis2016generativechoreographyusingdeep} extended this work with mixture density networks (MDNs), reducing the risk of averaging artifacts. However, they used only a single front camera for motion capture, limiting the recording of 3D movement data.\\
Other works leveraged latent space explorations, for example \cite{10.1007/978-3-319-77583-8_17}. The paper used autoencoders and trained them on the fly for an interactive, performative piece. There, trajectories in the latent space of the encoder have been sampled and displayed as possible continuation of the incoming movement sequence. \\
\cite{pettee2019imitationgenerativevariationalchoreography} combined the efforts of previous works by leveraging RNNs and (variational) autoencoders. \cite{raab2023singlemotiondiffusion} utilizes a more modern architecture, Denoising Diffusion models. They introduced a local attention mechanism \cite{arar2022learnedqueriesefficientlocal} in the denoising process, which led the model to be able to perform various generative tasks trained on a single movement sequence for an arbitrary skeleton representation. While they utilize successfully the style transfer capabilities of ILVR in order to harmonize example movement sequence to the single movement sequence they have trained their model on, we generalize this method to larger architectures and datasets, and combine it with time-coherent motion inpainting to obtain an interactive system.

\paragraph{Music-conditioned generation}
Combining different modalities has sparked interest in the research community early on \cite{multimodaldeeplearning}. Having been always intertwined as ''create'' disciplines, music and dance offer a fertile ground to explore conditional generation from sound to movement sequences. \cite{inproceedingsgroovenet} have done such an exploration with Factored Conditional Restricted
Boltzmann Machines (FCRBM) and Recurrent Neural Networks
(RNN), but report poor generalization capabilities.  \\
More powerful and popular models are Bailando++ \cite{siyao2022bailando3ddancegeneration} and EDGE \cite{tseng2023edge}. Bailando++ consists of a Vector-Quantized Variational Autoencoder (VQ-VAE) and a Motion GPT predicting codebook motion sequences from music features. EDGE is a conditional denoising diffusion architecture, which utilized motion inpainting in order to create arbitrarily long dance sequences. Both have been trained and evaluated on the largest music and dance dataset AIST++ \cite{li2021ai}.\\
Most recent, notable pursuits incorporate foundation models \cite{liu2025dgfmbodydancegeneration} or a mixture of several modalities \cite{zhang2025motionanythingmotiongeneration}, even the reverse: motion-to-music generation \cite{reimagingdance}.

\paragraph{Duet dance interaction}
While the previous models build on single-person data, recent attempts have been made to create comparable, large scale multi-person dance datasets and specialized models. 
\cite{siyao2024duolandofollowergptoffpolicy} presents the first large-scale duet dataset and train a VQ-VAE + GPT structure on it. \cite{li2024interdancereactive3ddancegeneration} introduces another large-scale duet dataset with a diffusion model whose conditional guidance process has been refined in order to avoid unphysical results like body surface penetrations. Finally, \cite{ghosh2025duetgenmusicdriventwoperson} introduce a hierarchical VQ-VAE \cite{takida2024hqvaehierarchicaldiscreterepresentation} model which outperforms both on several metrics. However, the open question whether models trained on human-human datasets generalize to human-AI interaction tasks remains non-trivial. We circumvent this problem by offering a different perspective, utilizing features the model extracts from single-motion sequences to form an interactive system.




\section{Methods}
In this section, we introduce the methods used to build the interaction model. First, we discuss the backbone of the generative part: denoising diffusion models. It is followed by the motion inpainting algorithm which leads to an iterative, time-coherent motion continuation. Finally, the chosen style transfer method is specified and how all together leads to an interaction algorithm applied during inference.

\subsection{Diffusion Model}
To learn motion sequences, we closely follow the implementation of EDGE \cite{tseng2023edge}. Their model is a conditional diffusion model that incorporates frozen Jukebox-encoded audio features \cite{dhariwal2020jukeboxgenerativemodelmusic} into the decoding process. Since our focus is movement generation with interaction, we omit these conditional aspects.

Denoising diffusion models leverage an iterative noising process in the forward pass. Let $\{\beta_t\}_{t=1}^T$ be a variance schedule, $\alpha_t = 1-\beta_t$, and $\bar{\alpha}_t = \prod_{s=1}^t \alpha_s$. The forward diffusion is
\begin{align}
q(\bm{x}_t \mid \bm{x}_{t-1}) &= \mathcal{N}\!\left(\bm{x}_t;\, \sqrt{\alpha_t}\,\bm{x}_{t-1},\, \beta_t \mathbf{I} \right), \label{eq:forward_markov}\\
q(\bm{x}_t \mid \bm{x}_0) &= \mathcal{N}\!\left(\bm{x}_t;\, \sqrt{\bar{\alpha}_t}\,\bm{x}_0,\, (1-\bar{\alpha}_t)\mathbf{I}\right). \label{eq:forward_closed}
\end{align}
A denoising network $\bm{\epsilon}_\theta(\bm{x}_t,t)$ predicts the noise, from which an estimate of the clean sample is made:
\begin{equation}
\hat{\bm{x}}_0(\bm{x}_t,t) \;=\; \frac{1}{\sqrt{\bar{\alpha}_t}}\,\bm{x}_t \;-\; \frac{\sqrt{1-\bar{\alpha}_t}}{\sqrt{\bar{\alpha}_t}}\,\bm{\epsilon}_\theta(\bm{x}_t,t).
\end{equation}
We train with the simplified noise prediction objective.
\begin{equation}
\mathcal{L}_{\text{denoise}} \;=\; \mathbb{E}_{\bm{x}_0,\, t \sim \mathcal{U}\{1,\dots,T\},\, \bm{\epsilon}\sim \mathcal{N}(\mathbf{0},\mathbf{I})}
\Big[\, w_t \,\big\| \bm{\epsilon} - \bm{\epsilon}_\theta\big(\sqrt{\bar{\alpha}_t}\bm{x}_0 + \sqrt{1-\bar{\alpha}_t}\bm{\epsilon},\, t\big)\big\|_2^2 \,\Big],
\label{eq:eps_objective}
\end{equation}
with optional per-step weights $w_t$.

\subsection{Auxiliary Loss}
To enhance physical realism, we add the following auxiliary loss terms: velocity loss and foot contact loss. While the velocity loss is introduced to inhibit jittering, the foot contact loss inhibits the sliding on the floor.

\paragraph{Velocity loss.}
Let a motion sequence be $\{\bm{p}_k\}_{k=1}^{T_{\text{seq}}}$ (stacked joint positions; $\Delta t$ is the frame interval). Ground-truth and predicted finite-difference velocities are
\begin{equation}
\bm{v}_k \;=\; \bm{p}_{k}-\bm{p}_{k-1}, \qquad
\hat{\bm{v}}_k \;=\; \hat{\bm{p}}_{k}-\hat{\bm{p}}_{k-1}, \quad k=2,\dots,T_{\text{seq}}.
\end{equation}
The velocity-matching loss is
\begin{equation}
\mathcal{L}_{\text{vel}} 
\;=\;\sum_{k=2}^{T_{\text{seq}}}
\big\| \hat{\bm{v}}_k - \bm{v}_k \big\|_2^2.
\label{eq:vel_loss}
\end{equation}

\paragraph{Foot-contact loss.}
Let $\mathcal{F}$ be the set of foot joints and define the speeds per joint $s_{k,j}=\|\bm{v}_{k,j}\|_2$. We mark contact frames by a threshold $\tau>0$,
\begin{equation}
m_{k,j} \;=\; \mathbf{1}\!\left[s_{k,j} < \tau\right], \qquad j\in\mathcal{F}.
\end{equation}
We penalize the predicted motion of feet during contact:
\begin{equation}
\mathcal{L}_{\text{fc}}
\;=\;
\sum_{k=2}^{T_{\text{seq}}}\;\sum_{j\in\mathcal{F}} 
m_{k,j}\;\big\|\hat{\bm{v}}_{k,j}\big\|_2^2.
\label{eq:fc_loss}
\end{equation}

\paragraph{Total objective.}
\begin{equation}
\mathcal{L}_{\text{total}}
\;=\; \mathcal{L}_{\text{denoise}} + \lambda_{\text{vel}}\,\mathcal{L}_{\text{vel}} + \lambda_{\text{fc}}\,\mathcal{L}_{\text{fc}},
\label{eq:total_loss}
\end{equation}
with $\lambda_{\text{vel}}$ and $\lambda_{\text{fc}}$ taking the default values of the EDGE implementation.
\subsection{Motion Inpainting}
Following~\cite{tseng2023edge,tevet2022humanmotiondiffusionmodel}, we employ motion inpainting for temporally consistent continuation of the sequence.

Given two samples $x_1$ and $x_2$ of length $T$, our aim is to modify $x_2$ so that the first half of $x_2$ equals the second half of $x_1$, and the second half of $x_2$ is a meaningful and smooth continuation of its first half. To that end, both samples are encoded through the forward diffusion (noising) process into the latent space. During each denoising iteration, the first half of $x_{2,t}$ is set equal to the second half of $x_{1,t}$.

\subsection{Motion Style Transfer}
\begin{figure}[H] 
\vspace{-6pt}
    \centering
    \includegraphics[width=0.6\textwidth]{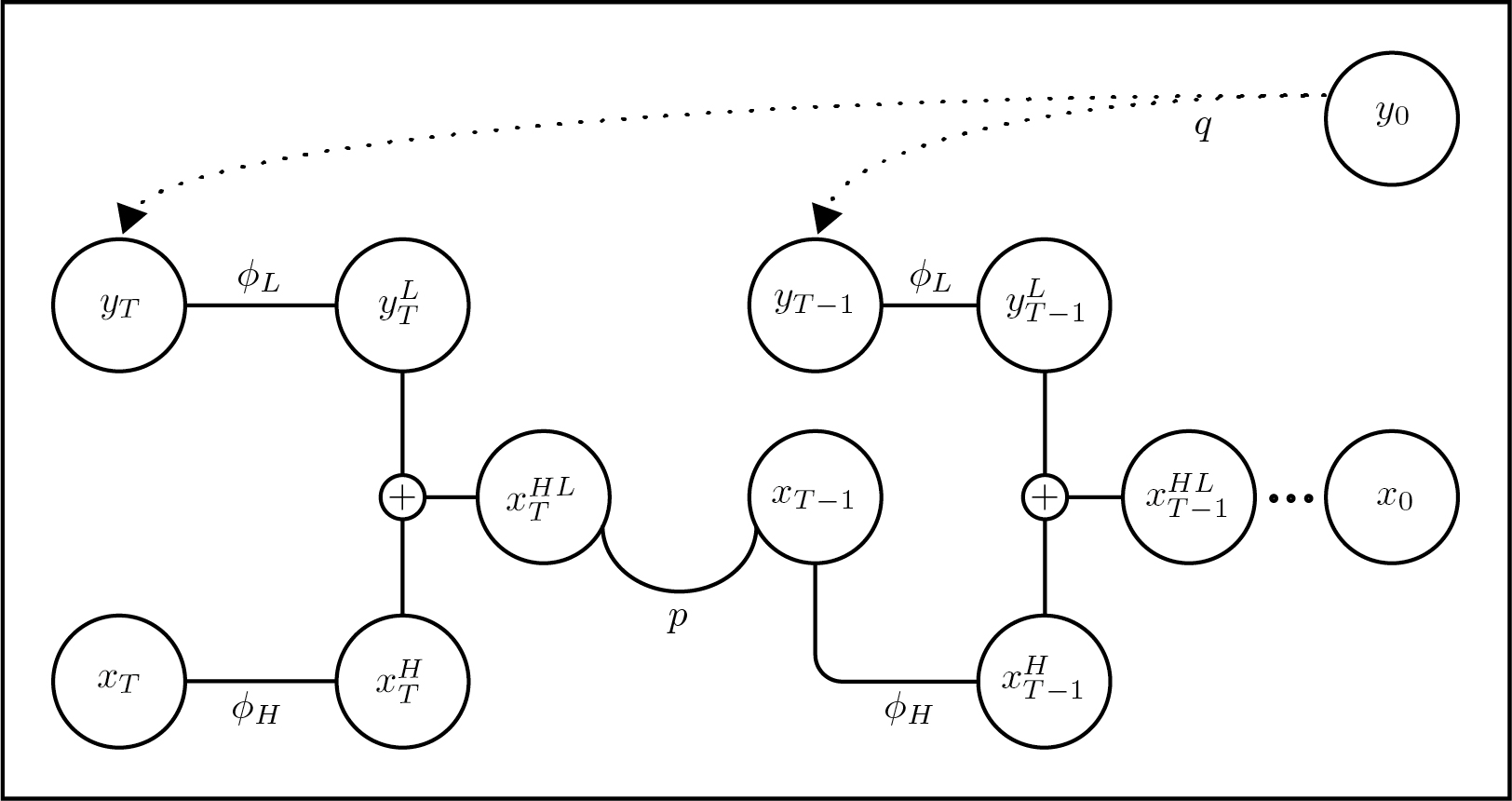}
    \vspace{-6pt}
    \caption{Depiction of the style-transfer process mimicking a reference sample $y_0$. The process starts with sampling $x_T$ from a normal distribution together with a noisy version of the reference sample $y_T$. In each iteration $t$, a low-pass filter $\phi_L$ is applied to $y_t$ and a high-pass filter $\phi_H$ to $x_t$. The sum is fed to the network $p$, which denoises it into the next iteration $x_{t-1}$.}
    \label{fig:style}
    \vspace{-6pt}
\end{figure}
Inspired by \cite{raab2023singlemotiondiffusion}, we use Iterative Latent Variable Refinement (ILVR) to mimic the motion of a reference sequence on the fly. Let $\phi_L$ be a low-pass operator (e.g., downsample\,$\rightarrow$\,upsample). We decompose a sample into low- and high-frequency components,
\begin{equation}
\bm{x} \;=\; \underbrace{\phi_L(\bm{x})}_{\text{low-frequency}} \;+\; \underbrace{\big(\bm{x}-\phi_L(\bm{x})\big)}_{\text{high-frequency}},
\end{equation}
and at each denoising step ($t{+}1 \!\to\! t$) replace the low-frequency component with that of a reference $\bm{x}^{\text{ref}}$:
\begin{equation}
\bm{x}_t \;\sim_p\; \phi_{HL}(\bm{x}_{t+1}) \;:=\; \phi_L(\bm{x_{t+1}}^{\text{ref}}) \;+\; \underbrace{\big(\bm{x}_{t+1}-\phi_L(\bm{x}_{t+1})\big)}_{\phi_H(\bm{x}_{t+1})}.
\end{equation}

Figure~\ref{fig:style} illustrates the iterative process of extracting the low-frequency components of the incoming performer sequence, denoted by $y_T$, through the filter $\phi_L$, and combining them with the high-frequency component of a random sample from the generative model, $X_T \sim \mathcal{N}(0, I)$, filtered through $\phi_H$, resulting in the new sample $x_T^{HL}$. The new sample is denoised into the next iteration of the backward process.

\subsection{Interaction}
\begin{figure}
\vspace{-6pt}
    \centering
    \includegraphics[width=0.6\textwidth]{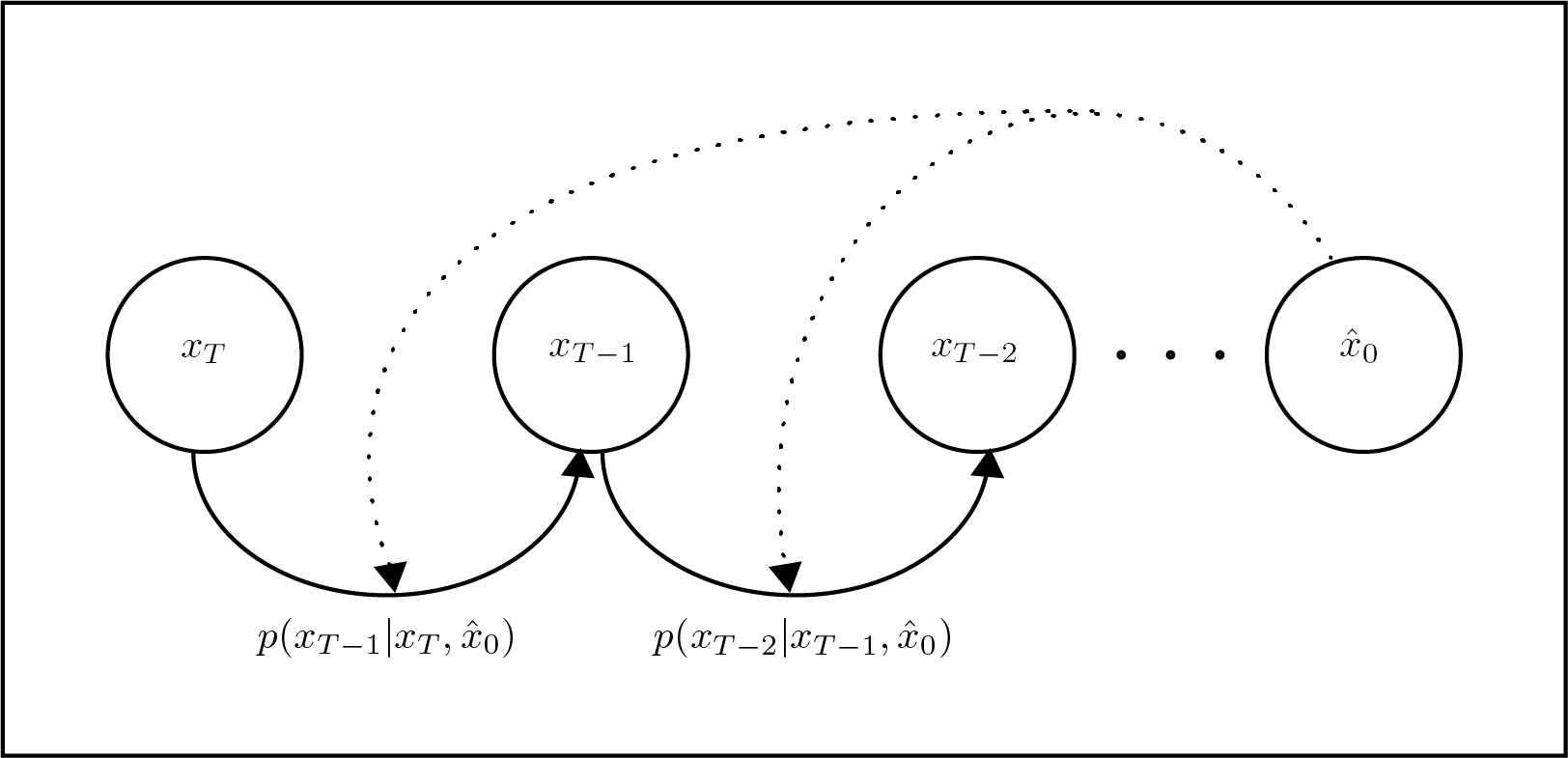}
    \vspace{-6pt}
    \caption{DDIM: Non-Markovian sampling process, predicting the denoised sample conditioned on the previous sample and an estimation of the clean version $\hat{x}_0$.}
    \label{fig:markov}
    \vspace{-6pt}
\end{figure}

To simulate interaction, we apply motion style transfer and inpainting consecutively to a random initial sample. The strength of the interaction is modulated by the denoising time step $T_\text{range}$, up to which the style-transfer function is applied. We use the non-Markovian denoising process of denoising implicit diffusion models (DDIM) \cite{song2022denoisingdiffusionimplicitmodels} to reduce the number of sampling steps—and therefore the inference time—by a factor of 20 without compromising generation quality. As illustrated in Figure~\ref{fig:markov}, at each denoising step $t$, the estimated clean sample $\hat{x}_0$ is fed into the posterior $p$ together with the previous sample $x_t$. This additional signal dramatically reduces the number of steps required compared to the original Markovian version.

The pseudo-code is given as Algorithm \ref{alg:long_ddim_sample_fixed}:

\begin{algorithm}
\caption{Long-DDIM Sampling with Temporal Stitching}
\label{alg:long_ddim_sample_fixed}
\begin{algorithmic}[1]

\State Initialize $x \sim \mathcal{N}(0, I)$;  
\For{\textbf{each} $(t_{i+1}, t_i) $} \Comment{$t_{i+1} < t_i$ for sampling}
  \State $(\hat{\epsilon}, \hat{x}_0) \gets \textsc{p}(x, \text{model}, t_{i+1})$
  \State $\alpha \gets \alpha_t$, \quad $\alpha' \gets \alpha_{t'}$
  \State $\sigma \gets \eta \cdot \sqrt{\frac{1-\alpha'}{1-\alpha}} \cdot \sqrt{1 - \frac{\alpha'}{\alpha}}$ \Comment{DDIM $\sigma$; if $\eta{=}0$, deterministic}
  \State $c \gets \sqrt{1 - \alpha' - \sigma^2}$
  \State $\xi \sim \mathcal{N}(0, I)$
  \State $x \gets \sqrt{\alpha'}\, \hat{x}_0 \;+\; c\, \hat{\epsilon} \;+\; \sigma\, \xi$ \Comment{DDIM step from $t_{i+1} \to t_i$}
  \If{\textit{with\_interaction}}
    \State $x \gets \phi^{HL}(x, t_i, \text{ref}_{sequence}, T_{\text{range}})$
  \EndIf
    \State $\tilde{x}_{\text{prev}} \gets \textsc{q}(x_{\text{prev}}, t_i)$ \Comment{$\tilde{x}_{\text{prev}}$ in the same noise level as $x$}
    \State $x[:, :half, :] \gets \tilde{x}_{\text{prev}}[:,half:, :]$ \Comment{copy 2nd half of prev into 1st half of current}
  
  \State $x_{\text{prev}} \gets x$
\EndFor
\State \textbf{return} $x$
\end{algorithmic}
\end{algorithm}

\section{Evaluation}
We evaluate by generating 3D motion sequences that enable interactive partner dancing, which is characterized by diversity of movements as well as alignment with a partner. In this work, we are primarily concerned with basic feasibility—i.e., generation quality—rather than factors such as generation speed. Thus, following prior work, we evaluate on a common dataset (AIST++ \cite{li2021ai}) and report quantitative metrics for diversity and alignment, alongside qualitative snapshots of generated sequences.

\begin{figure}[H] 
\vspace{-6pt}
    \centering
    \includegraphics[width=0.4\textwidth]{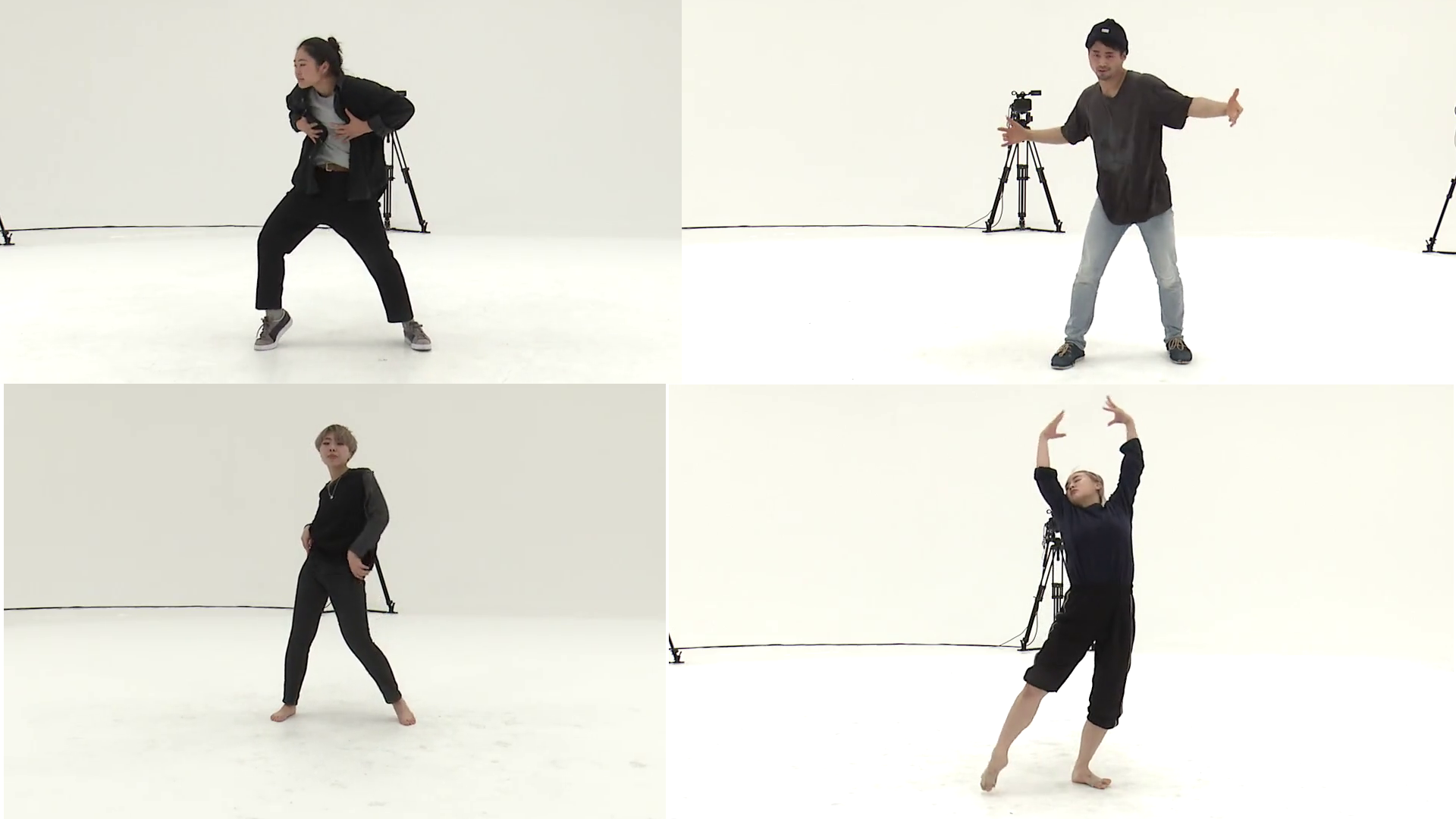}
    \vspace{-6pt}
    \caption{Snapshots from the music and dance dataset AIST++ \cite{li2021ai}. }
    \label{fig:aistpp}
    \vspace{-6pt}
\end{figure}
\paragraph{Dataset}
The largest dataset for standardized 3D dance movement data is the popular AIST++ dataset \cite{li2021ai}, which contains a diverse set of dance sequences across ten genres (illustrated in Figure \ref{fig:aistpp}). The training set contains seven genres, and the test set contains the remaining three, demonstrating the ability of our algorithm to generalize to unseen classes.

\paragraph{Implementation}
Since we omit the music-conditioning features, which connect music and motion representations, we replace cross-attention with self-attention and increase the number of layers from 8 to 16 to increase the model’s expressiveness. The interaction is controlled by the noise level (e.g., the maximum denoising step $T_{\text{range}}$) up to which style transfer is applied. The low-frequency filter $\phi_L$ is a consecutive down and up sampling through linear interpolation of factor 2, leading to smoothing out high frequency components.  We perform 50 DDIM denoising steps, following the implementation of EDGE.

\paragraph{Metrics}
To quantify the degree of mimicry, we use the Fréchet Inception Distance (FID) \cite{Guo_2020,heusel2018ganstrainedtimescaleupdate} and a diversity measure. FID is a standard and widely used evaluation metric in generative modeling, particularly for assessing the similarity between real and generated data distributions. Specifically, we compute the distributions of the kinetic energies of individual joints in the dataset and in the generated samples, and we measure the distances between these distributions.

\paragraph{Results}
\begin{figure}[H] 
\vspace{-6pt}
    \centering
    \includegraphics[width=0.6\textwidth]{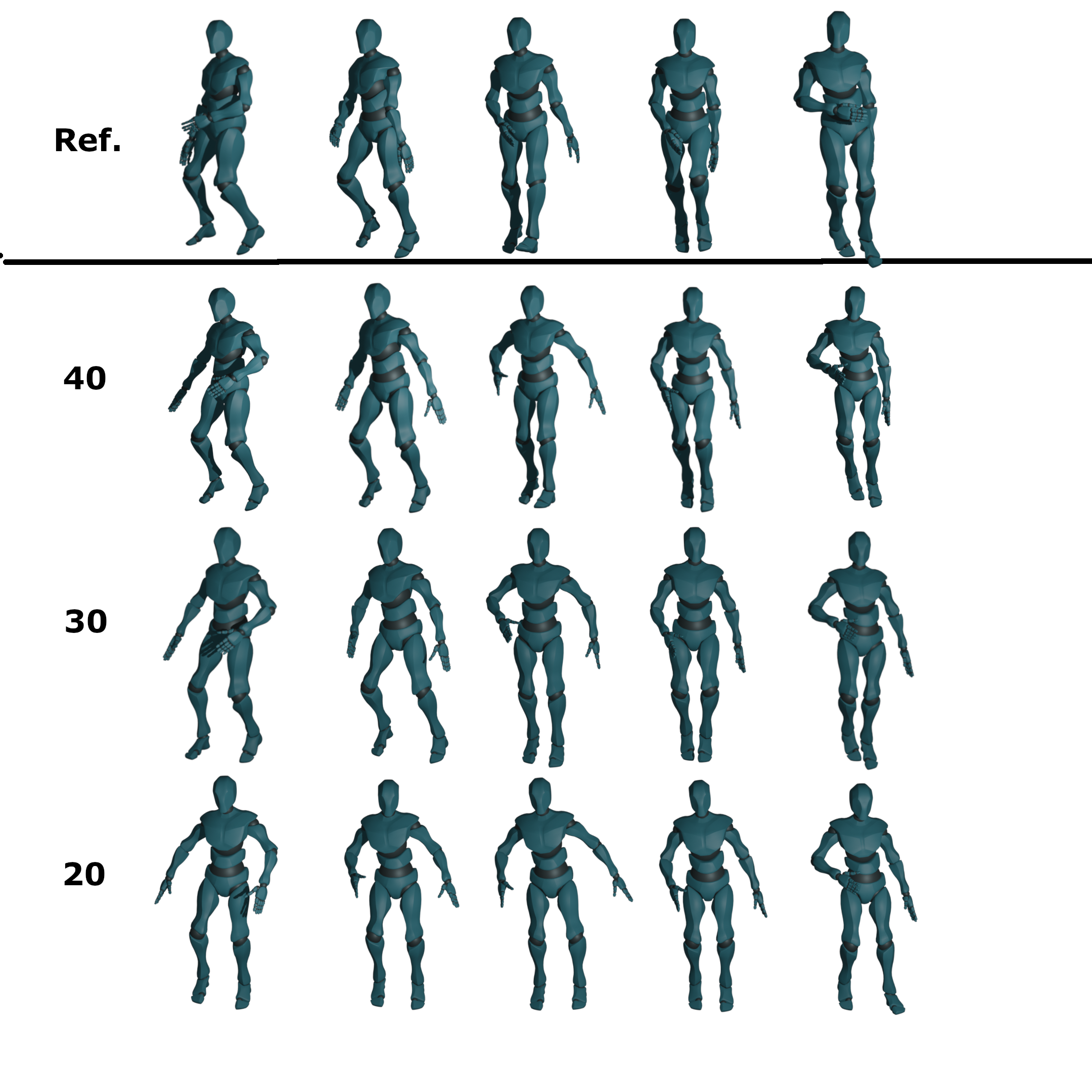}
    \vspace{-6pt}
    \caption{Snapshots of samples from the test set (first row) compared to snapshots of model with interaction strength 40,30,20 at same time frame. One can observe that increasing interaction strength leads to stronger mimicry, i.e. in the orientation of the body and the swinging of the arms.}
    \label{fig:all}
    \vspace{-6pt}
\end{figure}
In Table~\ref{tab:results}, random sampling from the unconditional EDGE model is compared with samples exhibiting varying interaction strengths. We observe that the longer the style transfer is applied during denoising, the closer the generated feature distribution is to the ground truth, as reflected in FID.

For diversity, one might expect the opposite: higher interaction strengths impose greater constraints on movement and should therefore reduce diversity. However, we observe that the unconditional EDGE model (no interaction) attains the lowest diversity score, suggesting that the base model’s generalization is weak. Consequently, mimicking the test set initially increases diversity before the expected decline.

Examining the diversity metrics more closely, we see an increase in diversity as interaction strength grows, followed by a slight decline at the highest strengths. Because the score is lowest for the unconditional EDGE model, it is plausible that the base model does not generalize well.

Figure~\ref{fig:all} illustrates the effect of different interaction strengths in producing mimicry for a given reference sequence. The top row displays snapshots from the reference movement; the following rows show samples from our model with decreasing interaction strength, controlled by $T_{\text{range}}$. Several aspects point to the effectiveness of this control parameter: the body orientation is maintained across the sequence for the $40$ setting, while it is only weakly correlated for the $20$ setting. The arm motion seems to undergo a similar transition as the reference movement as well, with $40$ exhibiting slightly stronger dynamics than $20$. This example indicates that style transfer—with a decomposition into low- and high-frequency components—both adapts to the reference sequence and modifies it creatively.

\begin{table}[h]
\centering
\caption{Kinetic feature metrics. ``Test'' is ground truth, ``Uncon. EDGE'' is the baseline, and 10--40 denote different interaction strengths. All distribution distance scores has been calculated with the combined training and test set as reference, as done in the evaluation of Bailando++. Best scores (except GT) are bold.}
\begin{tabular}{lcc}
\toprule
Setting & FID$_k$ $\downarrow$ & Div$_k$ $\uparrow$ \\
\midrule
Test (GT)   & 9.55  & 6.57 \\
Uncon. EDGE     & 111.95 & 2.64 \\
Interaction 10  & 105.50 & 2.84 \\
Interaction 20  & 97.34  & \textbf{3.89} \\
Interaction 30  & 72.89  & 3.65 \\
Interaction 40  & \textbf{49.14}  & 3.56 \\
\bottomrule
\end{tabular}
\label{tab:results}
\end{table}

\section{Conclusion and Future Work}
We leveraged motion style transfer and motion inpainting to create a time-coherent, interactive diffusion model based on single-person motion data. We showed that increasing the number of denoising steps at which the transfer is applied modulates convergence toward the reference movement. We also explored low-pass and high-pass filters for separating the content and style of a movement sequence. Future work may investigate other unsupervised editing methods in diffusion models (e.g., \cite{chen2024exploringlowdimensionalsubspacesdiffusion}) for richer interaction controls, as well as knowledge distillation \cite{meng2023distillationguideddiffusionmodels} or techniques that reduce the number of inference steps (as in \cite{dai2024motionlcmrealtimecontrollablemotion}, using consistency models \cite{song2023consistencymodels}) to approach real-time performance.

\subsection{Societal impact}
This work adds another modality to the artistic exploration of machine-learning algorithms as an artificial other. Alongside the success of large language models and early improvisational algorithms for music co-creation, this work offers a first attempt to utilize high-level features learned from single-person motion data for interactive purposes. In a similar vein, the authors of \cite{f4e54e32c7b148c981febc9abbe3ab3b} explore how glitches in movement generation—something that might initially be considered a limitation of the model—can be used for inspiration by relating them to non-physical movement.\\
In the spirit of \cite{Robbins2023}, who argues that artistic thinking, as a creative framework for innovation, can be applied effectively beyond the arts (e.g., in management and R\&D), we believe that making generative-AI frameworks accessible for artistic exploration can inspire new ideas and foster a different relationship to questions about their capabilities for creativity and intelligence.\\
Furthermore, we envision that, in the future, our work could contribute to well-being by enabling people to practice movement freely with an AI partner—an entity available 24/7, free from expectations of a partner and social pressure. Ultimately, we see human–AI dance as a complement to, rather than a replacement for, human–human dancing, potentially opening new forms of creative and embodied interaction.

\subsection*{GenAI Usage Disclosure}
In this work, a Large Language Model was used to complement literature search in combination with classical search on Google Scholar including forward/backward search.
\bibliographystyle{plainnat}
\bibliography{iclr2026_conference}

\end{document}